\documentclass[10pt,twocolumn,letterpaper]{article}
\usepackage[pagenumbers]{wacv}
\usepackage{microtype}
\usepackage{booktabs}
\usepackage{enumitem}
\usepackage{pifont}
\usepackage{xspace}
\usepackage{xcolor}
\usepackage{amsmath,amssymb}
\usepackage{graphicx}
\usepackage{placeins}

\newcommand{\method}{VTC-AUDIT\xspace}
\newcommand{\onevision}{LLaVA-OneVision-2-8B-Instruct\xspace}
\newcommand{\onevisiontable}{\shortstack[l]{LLaVA-OneVision-2-\\8B-Instruct}}
\newcommand{\scfr}{\mathrm{SCFR}}
\newcommand{\cln}{\mathrm{cln}}
\newcommand{\dyn}{\mathrm{dyn}}
\newcommand{\frz}{\mathrm{frz}}
\newcommand{\cfa}{\mathrm{cf}}
\newcommand{\ful}{\mathrm{full}}

\setlist[itemize]{leftmargin=*,nosep}
\AtBeginDocument{\hfuzz=0pt\vfuzz=0pt}

\definecolor{wacvblue}{rgb}{0.21,0.49,0.74}
\usepackage[breaklinks,colorlinks,allcolors=wacvblue]{hyperref}
\hypersetup{pdfauthor={Isaac(Rucheng) Li},pdftitle={VTC-AUDIT: An Interventional Reliability Audit for Visual-Token Compression under Paired Image Shifts}}
\usepackage{placeins}

\title{\method: An Interventional Reliability Audit for Visual-Token Compression under Paired Image Shifts}
\author{Isaac(Rucheng) Li\\University of Pittsburgh\\\href{mailto:isl74@pitt.edu}{\texttt{isl74@pitt.edu}}}
\date{}

\begin{document}
\maketitle

\begin{abstract}
Visual-token compression is usually evaluated by clean accuracy and
efficiency, but these averages do not reveal failures that compression
introduces under an image shift where the uncompressed model stays
correct. We introduce \method, which estimates this incremental risk
only on pairs where the full model is correct on both images and the
compressed model is correct on the clean image. The four arms then
cross two binary choices: which image selects the retained token
positions, and which image supplies the visual features. We audit three
open 8--9B vision-language models on four tasks and eight pre-specified
shifts. We evaluate two training-free compressors---hard-pruning
DivPrune and prune-and-merge ApET---at a matched 25\% retained-position
budget, producing 708,000 inference records. Reselecting positions on
the shifted image, rather than reusing the clean-image positions, raises
the failure rate in all six model--compressor settings. The increase is
3.04--9.60 percentage points for DivPrune and 2.05--4.20 percentage
points for ApET, with every 95\% interval above zero. Repeating the
audit on a separately resampled second manifest reproduces all six
increases. In a pooled monitor audit across the four benchmarks,
support Jaccard instability fails the practical AUROC criterion of
0.65 in all six settings, whereas clean compressed-model uncertainty
clears it. Thus, support turnover is a weak ranking signal under the
audited benchmark mixture, while clean compressed-model uncertainty
is substantially more informative. These results suggest that future
visual-token compression evaluations should report conditional
reliability alongside clean accuracy and efficiency.
\end{abstract}

\section{Introduction}

Large vision-language models (VLMs) process images as sequences of visual
tokens, often hundreds or thousands per input. Training-free visual-token
compression reduces this cost while preserving much of clean-task accuracy
\cite{alvar2025divprune,ma2026apet,zhang2025cdpruner}. However, clean accuracy and efficiency do not reveal whether compression creates new failures under image shifts that the uncompressed model still handles correctly.

We condition on pairs where the full model is correct on both images and
compression succeeds on the clean image, defining the population on which
strict compression-specific failure rate (SCFR) is measured.

Image shifts can affect both visual features and which visual-token indices a
compressor retains. We refer to the set of retained indices as the
\emph{support}. \method therefore forms a four-arm intervention by crossing the
image used to select the support with the image supplying the visual features.
The dynamic--frozen contrast measures the additional failure from selecting
support on the shifted image rather than reusing the clean-image support, while
the clean-feature counterfactual tests whether shift-selected support remains
harmful when applied to clean features. We also ask whether larger changes in
the retained-index set are more predictive of failure, using Jaccard
instability and clean compressed-model uncertainty as two simple predictors.

Our audit combines paired full-model conditioning, four-arm
support/feature interventions, and per-example failure prediction
under non-adversarial image shifts. We evaluate this protocol on two
text-agnostic compressors with different operators: DivPrune discards
unselected tokens, while ApET merges their information into the retained
positions.

Across three open 8--9B VLMs, four tasks, eight image shifts, and both
compressors, the audit shows a consistent four-arm pattern: dynamic and
clean-feature-counterfactual failure rates are close to each other and well above
frozen-support failure. Reselecting positions on the shifted image, rather than reusing the clean-image
positions, raises the failure rate in every model--compressor setting. Across
the three models the increase spans 3.04--9.60 percentage points for DivPrune
and 2.05--4.20 percentage points for ApET, every 95\% interval excludes zero, and a separately resampled second manifest reproduces all six increases. Moreover, 74.8--81.6\% of DivPrune dynamic failures
and 55.6--64.1\% of ApET dynamic failures persist after restoring clean
features, showing substantial but compressor-dependent support sensitivity.

Support movement itself is nevertheless a weak per-example failure signal.
On the primary seed-42 sample, pooled across the four benchmarks, the
orientation-free Jaccard AUROC bound is at most
0.605 for DivPrune and 0.592 for ApET, both below the pre-specified 0.65 target
on every model. Clean compressed-model uncertainty is substantially stronger,
with AUROC 0.705--0.844 across the two compressors. These results motivate
conditional reliability as a complement to clean accuracy and efficiency.

Code, manifests, and experimental outputs are available at
\url{https://github.com/IsaacLi74/VTC-AUDIT}.

\section{Related Work}

\paragraph{Visual-token compression.}
DivPrune greedily maximizes diversity among retained visual tokens
\cite{alvar2025divprune}; ApET ranks tokens by linear approximation error and
merges removed information into retained positions \cite{ma2026apet}. Both are text-agnostic: neither conditions selection on the user instruction. CDPruner instead conditions selection on the instruction via a determinantal
point process \cite{zhang2025cdpruner}, and is outside our text-agnostic scope.
These methods are primarily evaluated through clean accuracy--efficiency trade-offs.

\paragraph{Compression evaluation and failure analysis.}
VTC-Bench filters existing benchmarks for examples that better distinguish
visual-token compression methods \cite{liao2026vtcbench}, while LLMC+
standardizes comparisons across models, tasks, and compressors
\cite{lv2026llmcplus}. Acceleration-induced prediction instability under
input perturbations is examined in \cite{sun2025instability}. Wang et al.\ show that existing training-free pruning methods can perform no
better than random pruning in deeper VLM layers, attributing this degradation
to vanishing visual-token information \cite{wang2026informationhorizon}. \method instead asks a different
reliability question: when the full model remains correct, does compression
introduce a new failure under a non-adversarial image shift, and through which
support/feature pathway?

\paragraph{Compression-specific robustness.}
Compression-aware attacks show that token-ranking instability can induce failures
specific to compressed inference \cite{zhang2026pitfalls}, and that robustness
evaluated without compression is overly optimistic \cite{zhang2026cage}. A
concurrent provenance audit examines whether retained tokens remain spatially
traceable to supporting evidence \cite{liu2026provenance}. Compression-specific
adversarial failures and counterfactual support reallocation are studied in
\cite{li2026cira}. We study non-adversarial paired image shifts instead, conditioning on full-model
correctness and localizing compression-specific failures through
support/feature interventions.

\section{Audit Protocol}
\subsection{Notation}
A \emph{configuration} is a quadruple $(x,q,a,T)$: a clean image, its question,
the benchmark's reference answer, and a shift $T$. We use four transformation
families, each at two severities (Sec.~\ref{sec:design}), so $T$ ranges over
eight shifts. Write $x'=Tx$ for the shifted image. The triple $(x,q,a)$ is a
single benchmark record and is held fixed across every arm; only the visual side
varies.

Model $u$ factors into a visual stage $\phi_u$ and a language stage $g_u$.
Because the question is fixed across all arms, we suppress it, together with
model-specific positional metadata, from the notation. For a given
configuration, write
\[
V_u=\phi_u(x)\in\mathbb{R}^{N_u\times d_u},\qquad
V'_u=\phi_u(x')\in\mathbb{R}^{N_u\times d_u},
\]
where $N_u$ is the common clean--shift visual-token count for the
configuration and $d_u$ is the model-specific feature dimension. Neither
depends on the compressor.

A compressor $m$ acts between the two stages of any model, and is a map on token
matrices alone. It comprises a selector
$S_m(V)\subseteq\{1,\dots,N_u\}$ returning $k<N_u$ retained token
indices, which we call the \emph{support}, and an emitter
$C^m_S(V)\in\mathbb{R}^{k\times d_u}$ returning the token matrix it produces from
$V$ under an arbitrary support $S$. The compressed answer is
$g_u\bigl(C^m_S(V)\bigr)$.

Each benchmark supplies a scoring rule $\sigma(\hat a)\in[0,1]$, the credit given
to an output $\hat a$ against the reference $a$; $\sigma=1$ is full credit and
$\sigma=0$ is zero credit. The rule is determined by the task, which is fixed
within a configuration. Three of our four tasks are effectively binary, while
TextVQA's soft consensus is graded, so $\sigma$ can take intermediate values.

\subsection{The four arms}
Every arm has the form $g_u\bigl(C^m_S(\cdot)\bigr)$. The arms differ in only
two respects: which image the support $S$ was computed from, and which image's
features it is applied to.
\begin{center}
\small
\begin{tabular}{llcc}
\toprule
Arm & Answer & Support & Features\\
\midrule
Clean          & $\hat a^{u,m}_{\cln}$ & $S_m(V_u)$  & $V_u$\\
Dynamic        & $\hat a^{u,m}_{\dyn}$ & $S_m(V'_u)$ & $V'_u$\\
Frozen         & $\hat a^{u,m}_{\frz}$ & $S_m(V_u)$  & $V'_u$\\
Counterfactual & $\hat a^{u,m}_{\cfa}$ & $S_m(V'_u)$ & $V_u$\\
\bottomrule
\end{tabular}
\end{center}
Clean and dynamic are the two paths a deployed system actually takes. The other
two are hypothetical: frozen withholds reselection, and counterfactual applies
the shift-selected support without feature drift. We write
$\hat a^u_{\ful}=g_u(V_u)$ and $\hat a'^u_{\ful}=g_u(V'_u)$ for the uncompressed
answers for model $u$, which carry no compressor index.

\subsection{Eligibility and strict failure}
\label{sec:eligibility}
For model $u$ and compressor $m$, a configuration is eligible when
\[
E^{u,m}=\mathbf{1}\bigl[\sigma(\hat a^u_{\ful})=\sigma(\hat a'^u_{\ful})
=\sigma(\hat a^{u,m}_{\cln})=1\bigr].
\]
Thus the uncompressed model must receive full credit on both the clean
and shifted images, and compression must receive full credit on the
clean image. This conditioning defines the target population of the
audit: configurations for which the underlying VLM remains correct
under the shift and clean compressed inference succeeds. SCFR therefore
characterizes compression-specific reliability after successful clean
compressed inference, rather than the compressed model's overall error
rate. Because eligibility includes clean compressed success under
$S_m(V_u)$, all arm rates and contrasts are interpreted within this
conditional population. Eligibility is compressor-specific through the
clean compressed prediction, so we report its coverage for every
model--compressor setting.

We compute rates separately in each of the 16 task--shift-family cells \(c\) (four tasks \(\times\) four shift families), with \(\mathcal{E}_c\) denoting the eligible configurations in cell \(c\). Dynamic SCFR is

\[
\scfr^{u,m}_{\dyn}=\operatorname*{mean}_{c}\;
\frac{\bigl|\{i\in\mathcal{E}_c:\ \sigma(\hat a^{u,m}_{\dyn,i})=0\}\bigr|}
     {\lvert\mathcal{E}_c\rvert},
\]

and frozen and counterfactual SCFR are defined analogously.

Within this eligible population, we define a strict failure as zero
credit on the arm being evaluated; partial-credit outputs remain in the
denominator but are not counted as failures. As a sensitivity analysis,
we also measure conditional score loss, $1-\sigma(\hat{a})$, using the
same eligible examples and bootstrap procedure.

Each task--shift-family cell includes both severities; we compute its rate over
all eligible configurations in that cell and then average the 16 cell rates
equally. Severity is analyzed separately; all arm rates and contrasts for a
model--compressor pair use the same eligible population.

\subsection{Contrasts}
We define the paired reselection contrast as
\[
\Delta_{u,m}=\scfr^{u,m}_{\dyn}-\scfr^{u,m}_{\frz}.
\]
The dynamic--frozen contrast keeps the shifted visual features $V'_u$
fixed and changes only which image is used to select the retained
support. Thus, $\Delta_{u,m}$ is the paired conditional contrast between
reselecting support on the shifted image and reusing the clean-image
support within the eligible population. The counterfactual arm tests
whether shift-selected support remains harmful when applied to clean
features $V_u$.

We also report persistence: among eligible dynamic failures, the fraction
that also fail in the counterfactual arm. For DivPrune, changing $S$
changes only which positions are retained. For ApET, it can also change
which removed tokens are merged into those positions. Accordingly, for
ApET, $\Delta_{u,m}$ characterizes the reselection pathway jointly
through the retained positions and the merge pattern induced by that
selection.

We obtain 95\% percentile intervals for \(\Delta_{u,m}\) from 5,000 image-cluster bootstrap resamples stratified by benchmark, keeping all shifts of a sampled image together.

\subsection{Monitor audit}
The four-arm contrast asks whether support reselection increases failure on
average. We next ask whether the amount of support movement predicts which
eligible pairs will fail. The binary label is dynamic failure,
$\sigma(\hat a^{u,m}_{\dyn})=0$, and the candidate score is Jaccard
instability,
\[
I_{u,m}
=
1-\frac{\lvert S_m(V_u)\cap S_m(V'_u)\rvert}
        {\lvert S_m(V_u)\cup S_m(V'_u)\rvert}.
\]
We evaluate ranking ability by AUROC over the pooled eligible
configurations from all four benchmarks, using the same
benchmark-stratified image-cluster bootstrap for 95\% intervals. As a
model-state comparator, we use clean compressed-model uncertainty,
defined as the negative mean token log-probability of the clean
compressed answer. The monitor therefore evaluates ranking performance
under the audited benchmark mixture, complementing the population-level
four-arm contrasts.

We apply the same monitor analysis to both compressors, treating
DivPrune as the pre-specified confirmatory setting and ApET as a
descriptive replication. We require at least 100 positive image clusters
and an AUROC interval width of at most 0.15, and pre-specify 0.65 as the
practical AUROC target. This threshold is a practical decision criterion
rather than a universal AUROC standard, so we report the continuous
AUROC estimates and their full 95\% intervals alongside the
threshold-based conclusions. Jaccard instability has no assumed
direction, so for a raw AUROC interval $[L,H]$ we use the conservative
orientation-free upper bound
$U_{\pm}=\max(H,1-L)$. We consider Jaccard practically uninformative
when $U_{\pm}<0.65$. For clean uncertainty, the direction is
pre-specified---higher uncertainty indicates higher risk---so it clears
the target only when its lower 95\% bound exceeds 0.65.

\section{Experimental Design}
\label{sec:design}
\paragraph{Models and tasks.}
We audit Qwen3.5-9B \cite{qwen35}, \onevision \cite{llavaonevision2}\footnote{HF revision:\linebreak\href{https://huggingface.co/lmms-lab-encoder/LLaVA-OneVision-2-8B-Instruct/tree/802f16c8346a062fef7ddf92b2bd64a770a50a1a}{\nolinkurl{802f16c8346a062fef7ddf92b2bd64a770a50a1a}}.}, and InternVL3.5-8B \cite{wang2025internvl35}. Each benchmark manifest contains 500 unique images from each of ChartQA \cite{masry2022chartqa}, TextVQA \cite{singh2019textvqa}, GQA \cite{hudson2019gqa}, and adversarial POPE \cite{li2023pope}, with one question per image. We use the benchmark scoring rules: normalized exact matching for GQA,
normalized yes/no matching for POPE, a 5\% relative tolerance for numeric
ChartQA answers (exact matching otherwise), and VQA consensus scoring for
TextVQA.

\paragraph{Paired shifts.}
We use four transformation families at two pre-specified
severities: JPEG quality 70/50, additive Gaussian noise $\sigma=2/5$ on the
0--255 scale, Gaussian blur radius 0.5/1.0, and brightness multipliers
0.9/0.8. These transformations preserve the post-projector visual-token grid,
providing the common index space required to exchange retained supports between
the clean and shifted representations in the frozen and counterfactual arms.
We verify equality of the token grids for every clean--shift pair.

Seed 42 defines the main benchmark sample; seed 2026 independently resamples
benchmark examples and Gaussian-noise realizations as a sensitivity check. The question, prompt formatting, and all non-visual model inputs are held fixed across arms. Decoding is greedy and deterministic throughout, so arm differences are not induced by sampling variability.

\subsection{The two audited compressors}
\label{sec:compressors}
Both compressors act on the same post-projector visual-token matrix
$V\in\mathbb{R}^{N\times d}$ consumed by the language model. We fix the
retained-position ratio at $r=k/N=0.25$, with
$k=\operatorname{round}(0.25N)$, defining a common compression operating point
across compressors. The budget fixes the number of retained positions, while
the algorithms determine \emph{which} positions survive and what information
those positions carry.

\paragraph{DivPrune: hard pruning.}
Let $V=[v_1,\ldots,v_N]^\top\in\mathbb{R}^{N\times d}$ denote the visual-token
matrix. DivPrune formulates token selection as a max--min diversity problem
\cite{alvar2025divprune}: for a target of $k$ retained positions, it seeks
\[
S^\star
=
\arg\max_{\substack{S\subseteq\{1,\ldots,N\}\\ |S|=k}}
\min_{\substack{i,j\in S\\ i\neq j}}
\bigl(1-\cos(v_i,v_j)\bigr).
\]
The method constructs this support greedily. It first selects the token whose
nearest neighbor is farthest away. Given the current support $S$, each
subsequent token is chosen as
\[
i^\star
=
\arg\max_{i\notin S}
\min_{j\in S}
\bigl(1-\cos(v_i,v_j)\bigr).
\]
Once $k$ positions are selected, all remaining tokens
are discarded and the retained tokens are restored to their original sequence
order before language-model inference.

\paragraph{ApET: approximation-error prune-and-merge.}
Let
$V=[v_1,\ldots,v_N]^\top\in\mathbb{R}^{N\times d}$
denote the visual-token matrix. ApET first uses farthest-point sampling to
select, by default, $M=10$ representative basis tokens, collected as
$B\in\mathbb{R}^{M\times d}$ \cite{ma2026apet}. It then measures how well
each visual token can be approximated by a linear combination of this small
representative subset.

Specifically, ApET seeks coefficients
$A\in\mathbb{R}^{N\times M}$ minimizing the total reconstruction error,
\[
A^\star
=
\arg\min_A \lVert V-AB\rVert_F^2.
\]
The normal-equation solution is
$A^\star=VB^\top(BB^\top)^{-1}$ when $BB^\top$ is nonsingular. In our
implementation, we use ridge stabilization with $\varepsilon=10^{-5}$:
\[
A=VB^\top(BB^\top+\varepsilon I)^{-1},
\qquad
\widehat V=AB.
\]
The approximation error for token $i$ is
\[
\xi_i=\lVert v_i-\hat v_i\rVert_2,
\]
where $\hat v_i$ is the $i$th row of $\widehat V$. For a target of $k$
retained positions, ApET keeps all $M$ basis tokens together with the
$k-M$ remaining tokens having the largest approximation errors. The remaining tokens are merged rather than simply discarded. Each removed token $j\notin S$ is assigned to its most cosine-similar retained token,
$\pi(j)=\arg\max_{i\in S}\cos(v_j,v_i)$. Writing $G_i=\{i\}\cup\{j\notin S:\pi(j)=i\}$ for the group assigned to retained
position $i$, ApET replaces $v_i$ by the group average
\[
\widetilde v_i
=
\frac{1}{|G_i|}
\sum_{j\in G_i} v_j.
\]

\paragraph{Audit-specific ApET instantiation.}
ApET originally applies compression at two stages: after the visual encoder
and again inside the LLM \cite{ma2026apet}. We use only the post-projector stage, where clean and shifted representations
have aligned visual-token grids, so a support selected from one image can be
applied to the corresponding token positions of the other. We retain ApET's
farthest-point sampling (FPS) for basis selection and its similarity-based
average merging. To make FPS deterministic, we choose its initial point using
a hash of the record UID, with the same initialization for each clean--shift
pair. In the frozen arm, the retained indices are fixed to those selected on
the clean image, while removed-to-retained merge assignments are recomputed
from the current arm's features. Because ApET merges information from removed
tokens whereas DivPrune discards them, the shared 25\% retained-position
budget does not imply equal retained information. We therefore treat
cross-compressor SCFR differences as descriptive rather than as a
head-to-head comparison of compressor quality.

\section{Results}

We report results for seed 42 unless noted otherwise; seed 2026 is used only
as a sensitivity check.

\begin{table*}[!t]
\centering
\caption{\textbf{Full-model performance, clean compressed performance, and
eligibility (seed 42).}
``Score'' is the mean benchmark score $\sigma$, and ``$\sigma=1$'' is the
percentage of outputs receiving full credit. Shifted full-model values pool
the eight shifts. Eligibility is the percentage of configurations satisfying
the conditions in Sec.~\ref{sec:eligibility}. All entries are percentages.}
\label{tab:benchmark_utility}
\scriptsize
\setlength{\tabcolsep}{2.6pt}
\resizebox{\textwidth}{!}{%
\begin{tabular}{llrrrrrrrrrr}
\toprule
& & \multicolumn{4}{c}{Full model}
& \multicolumn{3}{c}{DivPrune, 25\%}
& \multicolumn{3}{c}{ApET, 25\%} \\
\cmidrule(lr){3-6}\cmidrule(lr){7-9}\cmidrule(lr){10-12}
Model & Task
& Clean score & Clean $\sigma{=}1$ & Shifted score & Shifted $\sigma{=}1$
& Clean score & Clean $\sigma{=}1$ & Eligibility
& Clean score & Clean $\sigma{=}1$ & Eligibility \\
\midrule
Qwen3.5-9B & \textbf{All}
& 82.89 & 80.85 & 82.24 & 80.18
& 72.72 & 70.75 & 68.39
& 70.02 & 68.50 & 65.89 \\
& ChartQA
& 89.40 & 89.40 & 88.50 & 88.50
& 62.00 & 62.00 & 60.52
& 64.60 & 64.60 & 62.45 \\
& TextVQA
& 83.36 & 75.20 & 82.75 & 74.52
& 77.26 & 69.40 & 66.42
& 57.46 & 51.40 & 49.83 \\
& GQA
& 70.80 & 70.80 & 69.73 & 69.73
& 65.40 & 65.40 & 62.28
& 69.80 & 69.80 & 65.72 \\
& POPE
& 88.00 & 88.00 & 87.98 & 87.98
& 86.20 & 86.20 & 84.33
& 88.20 & 88.20 & 85.58 \\
\midrule
\onevisiontable & \textbf{All}
& 79.90 & 78.10 & 79.88 & 78.03
& 70.77 & 68.95 & 65.41
& 70.56 & 68.70 & 65.50 \\
& ChartQA
& 86.00 & 86.00 & 85.62 & 85.62
& 62.20 & 62.20 & 59.65
& 65.20 & 65.20 & 62.00 \\
& TextVQA
& 76.60 & 69.40 & 76.50 & 69.08
& 67.06 & 59.80 & 55.55
& 63.04 & 55.60 & 51.88 \\
& GQA
& 69.00 & 69.00 & 69.80 & 69.80
& 67.20 & 67.20 & 61.90
& 68.20 & 68.20 & 64.20 \\
& POPE
& 88.00 & 88.00 & 87.60 & 87.60
& 86.60 & 86.60 & 84.52
& 85.80 & 85.80 & 83.93 \\
\midrule
InternVL3.5-8B & \textbf{All}
& 74.14 & 72.70 & 73.83 & 72.39
& 56.40 & 55.25 & 51.41
& 53.08 & 52.30 & 49.26 \\
& ChartQA
& 78.00 & 78.00 & 76.55 & 76.55
& 29.80 & 29.80 & 27.10
& 33.00 & 33.00 & 29.88 \\
& TextVQA
& 65.36 & 59.60 & 65.00 & 59.23
& 46.62 & 42.00 & 39.38
& 30.72 & 27.60 & 25.72 \\
& GQA
& 69.20 & 69.20 & 69.38 & 69.38
& 66.20 & 66.20 & 59.95
& 65.60 & 65.60 & 60.77 \\
& POPE
& 84.00 & 84.00 & 84.40 & 84.40
& 83.00 & 83.00 & 79.22
& 83.00 & 83.00 & 80.67 \\
\bottomrule
\end{tabular}%
}
\end{table*}

\subsection{Performance and eligibility}

Table~\ref{tab:benchmark_utility} summarizes model performance and eligibility.
At the model-level aggregate, the shifts minimally affect full-model
performance: the largest clean-to-shifted mean-score change is 0.65 percentage
points. Compression is more task-dependent, preserving nearly all POPE and most GQA
performance, with ChartQA generally showing the largest degradation; on
InternVL, the clean ChartQA score falls from 78.00\% for the full model to
29.80\% with DivPrune and 33.00\% with ApET. The compressors also differ by task: ApET yields higher clean ChartQA scores
on all three models but lower TextVQA scores, especially for Qwen and
InternVL. Eligibility remains
substantial, ranging from 49.26\% to 68.39\% across model--compressor settings.

\begin{table*}[!t]
\centering
\caption{\textbf{Primary intervention results (seed 42).}
SCFR values and persistence are percentages.
$\Delta_{u,m}$ is the dynamic--frozen difference in percentage points.
Persistence is the fraction of dynamic failures that also fail in the
counterfactual arm. Contrasts are computed from unrounded rates.}
\label{tab:primary}
\footnotesize
\setlength{\tabcolsep}{4.5pt}
\begin{tabular}{llrrrrr}
\toprule
Model & Compressor & Dynamic SCFR & Frozen SCFR &
Counterfactual SCFR & $\Delta_{u,m}$ [95\% CI] & Persistence \\
\midrule
Qwen3.5-9B & DivPrune
& 7.12 & 4.08 & 8.10
& 3.04 [2.21, 3.85] & 77.07 \\
& ApET
& 5.77 & 3.72 & 5.14
& 2.05 [1.30, 2.82] & 55.58 \\
\midrule
\onevisiontable & DivPrune
& 5.65 & 2.13 & 5.50
& 3.52 [2.81, 4.25] & 74.82 \\
& ApET
& 4.99 & 2.52 & 4.38
& 2.46 [1.87, 3.08] & 58.76 \\
\midrule
InternVL3.5-8B & DivPrune
& 14.24 & 4.63 & 14.20
& 9.60 [8.18, 11.13] & 81.60 \\
& ApET
& 9.88 & 5.68 & 9.18
& 4.20 [3.15, 5.28] & 64.05 \\
\bottomrule
\end{tabular}
\end{table*}

\begin{table*}[!t]
\centering
\caption{\textbf{Partial-credit sensitivity of the reselection effect (seed 42).}
Score loss is $1-\sigma$ and is reported as a percentage, macro-averaged over
task--shift-family cells on the same eligible population as SCFR.
$\Delta_{\rm score}$ is dynamic minus frozen score loss in percentage points,
with 95\% benchmark-stratified image-cluster bootstrap intervals. Contrasts
are computed from unrounded rates. The TextVQA panel isolates the graded
benchmark.}
\label{tab:partial_credit}
\footnotesize
\setlength{\tabcolsep}{3.8pt}
\begin{tabular}{llrrr@{\hspace{10pt}}rrr}
\toprule
& &
\multicolumn{3}{c}{All tasks} &
\multicolumn{3}{c}{TextVQA only} \\
\cmidrule(lr){3-5}\cmidrule(lr){6-8}
Model & Compressor &
Dynamic & Frozen & $\Delta_{\rm score}$ [95\% CI] &
Dynamic & Frozen & $\Delta_{\rm score}$ [95\% CI] \\
\midrule
Qwen3.5-9B
& DivPrune & 7.23 & 4.20 & 3.04 [2.21, 3.88]
& 4.88 & 4.23 & 0.65 [$-$0.82, 2.10] \\
& ApET & 5.89 & 3.79 & 2.09 [1.33, 2.82]
& 10.40 & 5.62 & 4.78 [2.60, 7.02] \\
\midrule
\onevisiontable
& DivPrune & 5.82 & 2.24 & 3.58 [2.87, 4.32]
& 5.34 & 1.75 & 3.59 [2.32, 4.93] \\
& ApET & 5.08 & 2.59 & 2.49 [1.88, 3.13]
& 5.82 & 2.10 & 3.73 [2.28, 5.38] \\
\midrule
InternVL3.5-8B
& DivPrune & 14.39 & 4.66 & 9.73 [8.24, 11.23]
& 13.44 & 4.82 & 8.63 [6.17, 11.28] \\
& ApET & 10.15 & 5.90 & 4.26 [3.19, 5.37]
& 10.05 & 6.33 & 3.72 [1.62, 6.16] \\
\bottomrule
\end{tabular}
\end{table*}

\subsection{Reselecting support consistently increases SCFR}

Table~\ref{tab:primary} reports the three non-clean arms; the clean arm is
omitted because eligibility requires full credit on it. Dynamic SCFR exceeds frozen SCFR in all
six model--compressor settings, by 3.04--9.60 percentage points for DivPrune
and 2.05--4.20 percentage points for ApET, with every 95\% interval above zero. The largest effects occur on InternVL, where SCFR falls from 14.24\% to 4.63\% for DivPrune and from
9.88\% to 5.68\% for ApET.

Table~\ref{tab:partial_credit} examines the same reselection effect while
retaining partial benchmark credit. Unlike strict SCFR, this analysis uses score loss $1-\sigma$, reported as a
percentage, for every eligible example and reports the average dynamic and
frozen losses together with their difference $\Delta_{\rm score}$.
Across all tasks, dynamic loss exceeds frozen loss in all six
model--compressor settings, with every 95\% interval for
$\Delta_{\rm score}$ above zero. Because TextVQA is the only graded benchmark,
we also report it separately: all six point estimates are positive, and five
of six intervals exclude zero.

\subsection{Shift-selected support remains harmful on clean features}

Table~\ref{tab:primary} shows substantial configuration-level persistence:
74.8--81.6\% of DivPrune dynamic failures and 55.6--64.1\% of ApET dynamic
failures also fail in the counterfactual arm after clean features are restored.
Persistence is lower for ApET, meaning that a smaller fraction of its dynamic
failures remain failures after restoring clean features.

\begin{table*}[!t]
\centering
\caption{\textbf{Task--shift heterogeneity of the reselection contrast
(seed 42).} Each entry is
$\Delta_{u,m}=\mathrm{SCFR}_{\mathrm{dyn}}-\mathrm{SCFR}_{\mathrm{frz}}$
in percentage points, computed over both severities within each shift
family. The table reports all 96 model--compressor--task--shift-family
cells. Eligible cell sizes range from 256 to 857 configurations
(129--430 unique image clusters), with medians of 617 configurations
and 312 clusters. Cell-level contrasts characterize heterogeneity;
aggregate contrasts and their intervals are reported in Table~\ref{tab:primary}.}
\label{tab:task_shift}
\scriptsize
\setlength{\tabcolsep}{3.0pt}
\begin{tabular}{llrrrrrrrr}
\toprule
& & \multicolumn{4}{c}{DivPrune} & \multicolumn{4}{c}{ApET} \\
\cmidrule(lr){3-6}\cmidrule(lr){7-10}
Model & Task & JPEG & Noise & Blur & Bright. & JPEG & Noise & Blur & Bright. \\
\midrule
Qwen3.5-9B & ChartQA & 9.12 & 8.21 & 8.79 & 10.23 & 3.37 & 2.55 & 2.89 & 2.72 \\
 & TextVQA & 0.15 & 0.75 & 0.30 & 1.50 & 7.06 & 6.40 & 3.62 & 1.40 \\
 & GQA & 2.25 & 0.48 & 1.95 & 2.07 & 0.31 & 1.66 & $-$0.15 & 0.45 \\
 & POPE & 0.48 & 1.07 & 0.95 & 0.36 & $-$0.12 & 0.00 & 0.82 & $-$0.12 \\
\midrule
\onevisiontable & ChartQA & 9.76 & 8.51 & 8.43 & 10.33 & 6.16 & 3.04 & 7.64 & 4.17 \\
 & TextVQA & 3.83 & 2.85 & 3.45 & 3.21 & 4.50 & 2.68 & 5.61 & 1.71 \\
 & GQA & 0.99 & 0.96 & 1.13 & 1.44 & 0.47 & $-$0.62 & 2.04 & 0.31 \\
 & POPE & $-$0.36 & 0.12 & 0.71 & 0.94 & 0.84 & 0.48 & 0.12 & 0.24 \\
\midrule
InternVL3.5-8B & ChartQA & 28.25 & 23.27 & 19.17 & 31.75 & 9.36 & 11.84 & 12.07 & 11.59 \\
 & TextVQA & 7.36 & 8.31 & 9.49 & 7.36 & 3.49 & 3.49 & 3.12 & 3.89 \\
 & GQA & 2.86 & 2.49 & 1.69 & 5.11 & 1.99 & 2.62 & 1.32 & 0.82 \\
 & POPE & 1.27 & 2.64 & 0.51 & 2.13 & 0.00 & 0.00 & 0.62 & 0.99 \\
\bottomrule
\end{tabular}
\end{table*}

\subsection{The reselection effect is strongly task dependent}

The reselection effect varies sharply across tasks. In the pre-specified
DivPrune audit, Table~\ref{tab:task_shift} shows that ChartQA has the largest
positive $\Delta_{u,m}$ across all three models, while POPE is consistently
much less sensitive. Averaging first across the four shift-family cells within each model and then
across the three models, DivPrune has dynamic/frozen SCFR of
21.86\%/7.20\% on ChartQA, compared with 2.23\%/1.33\% on POPE.
ApET shows the same contrast, with 15.00\%/8.55\% on ChartQA versus
1.35\%/1.03\% on POPE. Qwen is an exception: its largest ApET reselection
effects occur on TextVQA rather than ChartQA.

The effect is also broad across transformations. Averaging equally across the 12 model--task combinations, $\Delta_{u,m}$ is
positive for JPEG, Gaussian noise, blur, and brightness under both compressors,
ranging from 4.71 to 6.37 percentage points for DivPrune and from 2.35 to
3.31 percentage points for ApET. Of the 48 task--shift-family cells per
compressor, 47 are positive for DivPrune; ApET has 41 positive, three zero,
and four negative cells. The negative cells are small in magnitude: the
minimum is $-0.62$ percentage points, compared with a maximum positive
contrast of 31.75 percentage points for InternVL--DivPrune on ChartQA under
brightness.

\subsection{Stronger shifts increase failure under fixed support}

Stronger shifts increase frozen SCFR in all six model--compressor settings
(Table~\ref{tab:secondary}), indicating greater failure under fixed
clean-image support.

Reselection remains harmful at both severities, but the dynamic--frozen gap
becomes smaller relative to dynamic SCFR. Using unrounded rates, the ratio
$\Delta_{u,m}/\scfr^{u,m}_{\dyn}$ decreases from mild to strong in all six
settings. On InternVL, it falls from 74.1\% to 60.6\% for DivPrune and from
48.3\% to 37.3\% for ApET.

\begin{table*}[!t]
\centering
\caption{\textbf{Shift-severity analysis (seed 42).}
For each model--compressor setting, the table reports dynamic SCFR, frozen
SCFR, and their difference $\Delta_{u,m}$ at mild and strong severity.
SCFR values are percentages; $\Delta_{u,m}$ is in percentage points.
Differences are computed from unrounded rates.}
\label{tab:secondary}
\footnotesize
\setlength{\tabcolsep}{5.0pt}
\begin{tabular}{llrrr}
\toprule
Model & Compressor &
Dynamic SCFR, mild $\rightarrow$ strong &
Frozen SCFR, mild $\rightarrow$ strong &
$\Delta_{u,m}$, mild $\rightarrow$ strong \\
\midrule
Qwen3.5-9B & DivPrune
& 6.67 $\rightarrow$ 7.58
& 3.47 $\rightarrow$ 4.69
& 3.20 $\rightarrow$ 2.88 \\
& ApET
& 5.30 $\rightarrow$ 6.25
& 3.40 $\rightarrow$ 4.04
& 1.90 $\rightarrow$ 2.21 \\
\midrule
\onevisiontable & DivPrune
& 5.34 $\rightarrow$ 5.95
& 1.84 $\rightarrow$ 2.42
& 3.51 $\rightarrow$ 3.53 \\
& ApET
& 4.40 $\rightarrow$ 5.58
& 2.21 $\rightarrow$ 2.84
& 2.19 $\rightarrow$ 2.74 \\
\midrule
InternVL3.5-8B & DivPrune
& 14.24 $\rightarrow$ 14.24
& 3.69 $\rightarrow$ 5.61
& 10.55 $\rightarrow$ 8.64 \\
& ApET
& 9.20 $\rightarrow$ 10.56
& 4.76 $\rightarrow$ 6.63
& 4.44 $\rightarrow$ 3.94 \\
\bottomrule
\end{tabular}
\end{table*}

\subsection{Support movement is a weak predictor of individual failure}

Support reselection consistently increases SCFR, but this
population-level effect does not imply that larger support changes
provide a useful ranking signal for dynamic-failure risk. We test this
directly with the pooled monitor audit in Table~\ref{tab:monitor}. All six
model--compressor settings satisfy the pre-specified sample-size and
precision requirements.

Jaccard instability is a weak pooled failure predictor. For DivPrune,
raw AUROC ranges from 0.423 to 0.492, and the orientation-free upper
bound $U_{\pm}$ is at most 0.605, below the 0.65 target on every model.
Thus, even allowing support movement to rank failure risk in either
direction, its discrimination does not reach the pre-specified
threshold. ApET shows the same pattern, with $U_{\pm}$ ranging from
0.535 to 0.592.

Clean compressed-model uncertainty performs substantially better in the
same pooled audit. Its AUROC ranges from 0.714 to 0.785 for DivPrune
and from 0.705 to 0.844 for ApET, with $L_{95}>0.65$ in all six
settings. The contrast is especially large on \onevision and InternVL3.5-8B
under DivPrune: Jaccard AUROC is 0.423 and 0.426, whereas uncertainty
reaches 0.783 and 0.785. The results therefore separate a
population-level reselection effect from per-example ranking under the
audited benchmark mixture: support turnover is a weak ranking signal,
whereas clean compressed-model uncertainty is substantially more
informative.

\begin{table*}[!t]
\centering
\caption{\textbf{Monitor audit (seed 42).}
AUROCs are reported with 95\% benchmark-stratified image-cluster bootstrap
intervals. $U_{\pm}$ is the orientation-free upper 95\% bound for Jaccard
instability, and $L_{95}$ is the lower 95\% bound for clean uncertainty.
The practical AUROC target is 0.65.}
\label{tab:monitor}
\scriptsize
\setlength{\tabcolsep}{4.0pt}
\begin{tabular}{llrrrr}
\toprule
Model & Compressor & Jaccard AUROC [95\% CI] & $U_{\pm}$ &
Uncertainty AUROC [95\% CI] & $L_{95}$ \\
\midrule
Qwen3.5-9B & DivPrune & 0.492 [0.465, 0.519] & 0.535 & 0.714 [0.678, 0.747] & 0.678 \\
& ApET              & 0.564 [0.537, 0.592] & 0.592 & 0.705 [0.663, 0.745] & 0.663 \\
\midrule
\onevisiontable & DivPrune & 0.423 [0.395, 0.453] & 0.605 & 0.783 [0.744, 0.815] & 0.744 \\
& ApET                  & 0.494 [0.465, 0.524] & 0.535 & 0.806 [0.778, 0.833] & 0.778 \\
\midrule
InternVL3.5-8B & DivPrune & 0.426 [0.395, 0.457] & 0.605 & 0.785 [0.754, 0.812] & 0.754 \\
& ApET                  & 0.488 [0.459, 0.517] & 0.541 & 0.844 [0.819, 0.867] & 0.819 \\
\bottomrule
\end{tabular}
\end{table*}

\subsection{Results reproduce on a second benchmark sample}

The separately resampled seed-2026 benchmark reproduces both main
conclusions. All six dynamic--frozen contrasts remain positive: the
increases span 3.09--8.71 percentage points for DivPrune and
2.23--4.76 percentage points for ApET, with every 95\% interval above
zero. The pooled monitor result is also unchanged: the largest
orientation-free Jaccard bound is 0.596, below the 0.65 target, while
the smallest lower 95\% bound for clean uncertainty is 0.654, above
the target.

\section{Discussion}

The results separate two questions about support sensitivity. Reselecting
support after an image shift consistently increases SCFR and score loss, yet
the amount of support turnover is a weak predictor of which individual
configurations fail. The population-level effect of reselection therefore does
not translate into a useful per-example signal based on set-level turnover
alone. This suggests that support sensitivity depends on more than the
magnitude of the support change.

The counterfactual arm provides complementary evidence about the role of
shift-selected support. A substantial fraction of dynamic failures remain
failures after clean visual features are restored, so these failures cannot be
attributed to shifted features alone. Persistence differs between the two
audited compression operators, with consistently lower values for ApET than
for DivPrune. More generally, the same amount of support change can have different effects
under different compression operators. DivPrune discards removed tokens,
whereas ApET merges their information into the retained positions, so support
overlap alone does not fully describe how the compressed representation changes.

These findings broaden what should be measured when evaluating visual-token
compression. The reselection effect appears across tasks and shift types, but
its magnitude varies substantially across tasks and with shift severity.
Clean performance alone therefore does not reveal whether compression
introduces new failures under input changes that the uncompressed model still
handles correctly. Reporting conditional reliability alongside accuracy and
efficiency makes this failure mode visible.

These findings suggest two practical directions. First, conditional
reliability should be characterized across compression budgets rather
than at a single operating point. Second, the dynamic--frozen gap
provides a concrete reliability target for support-stabilization and
adaptive-retention mechanisms; the frozen arm is diagnostic rather than
a proposed deployment policy. The weak performance of set-level Jaccard
further motivates predictors that distinguish which retained positions
change, rather than treating all support changes equally.

\section{Limitations}

\paragraph{Conditional estimand.}
Eligibility conditions on successful clean compressed inference under
$S_m(V_u)$, so the frozen arm reuses a support already known to succeed
on that configuration. Thus, $\Delta_{u,m}$ characterizes the
dynamic--frozen gap within this clean-successful population, but does
not separate shift-specific reselection from support replacement more
generally.

\paragraph{Empirical coverage and compression budget.}
Our empirical scope covers three open 8--9B VLMs, two training-free
text-agnostic compressors, four benchmarks, and four non-adversarial shift
families. The main audit uses a single 25\% retained-position operating point.
Exploratory 50\%-retention runs on a subset of configurations preserve the
qualitative direction of the main contrast, but we do not perform a systematic
sweep over retention ratios. The dependence of the reselection effect on
compression budget is therefore not characterized.

\paragraph{Shift validation and intervention applicability.}
We manually inspected 100 randomly sampled examples per task from the seed-42
benchmark sample to verify that the applied shifts preserved task semantics.
Nearly all inspected shifted examples remained human-answerable with the
original reference answer. However, this check covered only 100 of the 500
examples in each task and was not repeated for the independently resampled
seed-2026 benchmark sample. Semantic preservation is therefore not exhaustively
verified across the full evaluation set. The four-arm intervention also relies
on a shared clean--shift token index space. Geometry-changing transformations
or pre-projector compression settings without aligned token grids would require
an explicit correspondence between the two support spaces.

\paragraph{ApET instantiation.}
Our ApET audit uses only its post-projector compression stage, rather than the
original multi-stage operating point, to provide the aligned support space
required by the intervention.

\paragraph{Monitor scope.}
The monitor audit evaluates Jaccard support instability and clean
compressed-model uncertainty, with AUROC pooled across the four
benchmarks. It therefore characterizes ranking under the audited
benchmark mixture; task-specific monitor behavior is not analyzed
separately. We also do not evaluate alternative support-based failure
signals beyond set-level overlap.

\section{Conclusion}

We introduced \method, a paired interventional audit for evaluating
compression-specific reliability under image shifts. By conditioning on
cases where the uncompressed model remains correct and clean compressed
inference succeeds, the audit focuses on failures that emerge under
compressed inference despite preserved full-model correctness. Across
three VLMs and two training-free compressors at a 25\%
retained-position budget, reselecting support on the shifted image
consistently increases SCFR relative to reusing the clean-image support,
and many of these failures persist after clean visual features are
restored.

The audit also reveals a clear distinction between population-level
reliability and per-example ranking. Although support reselection
increases failure on average, support turnover is a weak ranking signal
in the pooled monitor audit, whereas clean compressed-model uncertainty
is substantially more informative under the same benchmark mixture.
Both conclusions reproduce on a separately resampled benchmark sample.
Together, these results show that clean accuracy and efficiency alone do
not capture the conditional reliability of visual-token compression and
motivate paired reliability auditing as a complement to conventional
compression evaluation.

{\small
\bibliographystyle{IEEEtran}
\bibliography{references}
}

\end{document}